%% file: main.tex
\documentclass{article}

\usepackage[preprint]{neurips_2022}

\usepackage[utf8]{inputenc} 
\usepackage[T1]{fontenc}    
\usepackage{hyperref}       
\usepackage{url}            
\usepackage{booktabs}       
\usepackage{amsfonts}       
\usepackage{nicefrac}       
\usepackage{microtype}      
\usepackage{xcolor}         

\usepackage{graphicx}
\usepackage{wrapfig,lipsum,booktabs}
\usepackage{multirow}
\usepackage[inline]{enumitem}

\definecolor{fig1pink}{RGB}{202, 57, 115}
\definecolor{fig1blue}{RGB}{17, 85, 204}

\title{Instruction Induction: From Few Examples\\to Natural Language Task Descriptions}

\author{Or Honovich$^\tau$ \qquad Uri Shaham$^\tau$ \qquad Samuel R. Bowman$^\nu$ \qquad Omer Levy$^\mu$$^\tau$\\
$^\tau$ Tel Aviv University\\
$^\nu$ New York University\\
$^\mu$ Meta AI\\
\texttt{\{or.honovich,uri.shaham1\}@gmail.com}}

\begin{document}

\maketitle

\input{00_abstract}
\input{01_intro}

\input{02_instruction_induction}
\input{03_data}
\input{04_evaluation}

\input{05_results}
\input{06_analysis}
\input{07_related_work}

\input{08_discussion}

\bibliographystyle{plainnat}
\bibliography{anthology,custom}


\appendix
\newpage
\input{09_appendix}

\end{document}

%% file: 00_abstract.tex
\begin{abstract}
Large language models are able to perform a task by conditioning on a few input-output demonstrations -- a paradigm known as \textit{in-context learning}.
We show that language models can explicitly infer an underlying task from a few demonstrations by prompting them to generate a natural language instruction that fits the examples.
To explore this ability, we introduce the \textit{instruction induction} challenge, compile a dataset consisting of 24 tasks, and define a novel evaluation metric based on \textit{executing} the generated instruction.
We discover that, to a large extent, the ability to generate instructions does indeed emerge when using a model that is both large enough and aligned to follow instructions;
InstructGPT achieves 65.7\% of human performance in our execution-based metric, while the original GPT-3 model reaches only 9.8\% of human performance.
This surprising result suggests that instruction induction might be a viable learning paradigm in and of itself, where instead of fitting a set of latent continuous parameters to the data, one searches for the best description in the natural language hypothesis space.\footnote{Our code and data are publicly available at \\\url{https://github.com/orhonovich/instruction-induction}}
\end{abstract}

%% file: 01_intro.tex
\section{Introduction}

Large language models (LMs) can perform unseen tasks by conditioning on a few labeled examples, effectively inferring the underlying tasks through a process known as \textit{in-context learning} \citep{gpt3}.
However, task inference is implicit, and the ability of models to \textit{explicitly} reason about it remains unexplored.
In this work, we show that LMs can explicitly describe an underlying task, in natural language, given a few labeled examples.

\begin{figure}[t!]
\centering
 \includegraphics[width=1.0\textwidth]{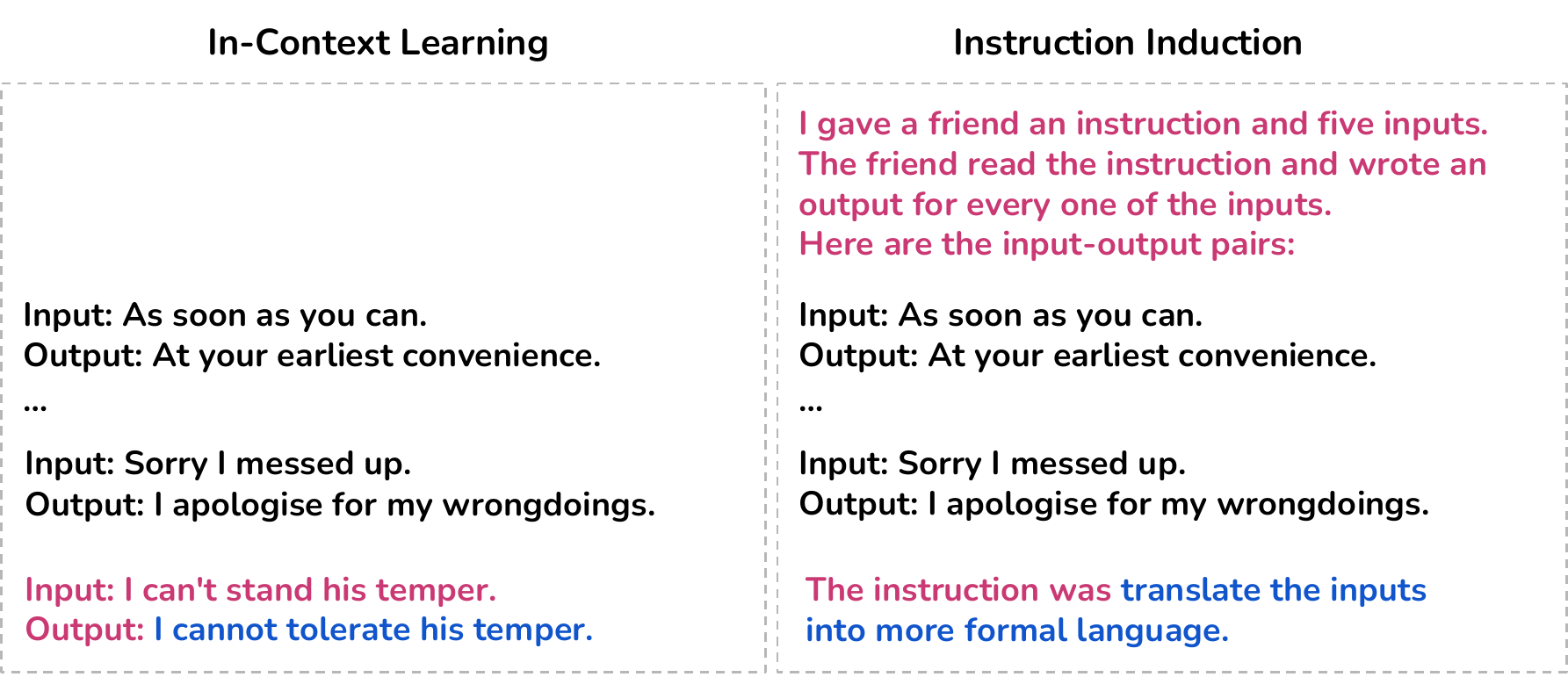}
 \caption{An example of instruction induction for the task of formality style transfer. \textit{Left:} the standard in-context learning setting; given five demonstrations, complete the sixth. \textit{Right:} instruction induction; the language model is prompted to generate a natural language instruction that describes the demonstrations. Model completions are in \textcolor{fig1blue}{blue}, prompt templates are in \textcolor{fig1pink}{pink}.}
 \label{fig:induction_example}
\end{figure}

We introduce the \textit{instruction induction} challenge, in which a model is provided with a few input-output demonstrations, and is requested to generate a natural language instruction describing the connection between the input-output pairs.
In our experiments, inducing instructions is done in a zero-shot manner by simply prompting the models to explain a small set of given demonstrations, as shown in Figure \ref{fig:induction_example};
we do not perform fine-tuning or use any labeled instruction induction data.

We examine instruction induction on 24 tasks, ranging from morphosyntactic tasks (e.g., pluralization) to style transfer (e.g., formality) and sentiment analysis.
As a basic evaluation protocol, we collect human annotations and use them as gold-standard references; the generated instructions are then compared to these references using BERTScore \citep{bertscore}.
Moreover, we suggest a novel evaluation metric for instruction induction: \textit{execution accuracy}. The execution accuracy of a generated instruction is measured by testing whether LMs can correctly perform the task in a zero-shot manner by using the generated instruction alone, without any demonstrations.

Our experiments reveal a surprising ability at generating correct instructions.
The best-performing model, InstructGPT \citep{instruct-gpt}, achieves an average BERTScore of 44.4, compared to human performance of 60.0; when measuring execution accuracy, the model reaches 43.6, with human-written instructions reaching 66.4.
For some tasks, the model's performance is on par or even better than human performance.
When qualitatively examining the generated instructions, we often observe accurate instructions, even for some of the more challenging tasks.
For instance, in the task of formality style transfer, generated instructions include ``Translate the inputs into more formal language'' and ``Use formal language''.
For semantic text similarity, the generated instructions include ``For each input, rate the similarity of the two sentences on a scale of 0 to 5, with 5 being a perfect match'' and ``Determine whether the two sentences are about the same thing''.

Despite these impressive results, we find that this ability is currently unique to InstructGPT \citep{instruct-gpt}, which is both very large (175B parameters) and was especially fine-tuned to follow instructions.
Ablations on smaller versions of InstructGPT as well as the original 175B-parameter GPT-3 \citep{gpt3} yield dramatically weaker performance.
These findings are in line with recent work showing that increasing model size unlocks new capabilities \citep{palm,predictability-and-surprise}, and serves as additional evidence for the strength of instruction tuning \citep{sanh2022multitask,wei2022finetuned,instruct-gpt}, perhaps even pointing to the necessity of complementing standard next-word prediction with additional objectives.


The fact that models can induce natural language instructions suggests that instruction-induction may serve as a learning paradigm of its own, where the goal is to find the best description in the natural language hypothesis space. While we currently provide a proof-of-concept for that idea, extending it by grounding models in natural language has the immediate benefit of human interpretability, and might also help alleviate overfitting and other issues associated with spurious correlations.

%% file: 02_instruction_induction.tex
\section{Instruction Induction}



We begin by formulating the task of instruction induction. Given a sequence of $n$ demonstrations $\{x_k, y_k\}_{k \in \{1,\ldots,n\}}$, the goal is to generate a \textit{single} natural language instruction, such that for each $x_k$, following the instruction results in $y_k$.
This format is similar to in-context learning \citep{gpt3}, only here the desired output is an \textit{instruction} describing the relation between the inputs and outputs of the demonstrations. We require models to perform this in a zero-shot setting, without any fine-tuning on labeled data.
Figure~\ref{fig:induction_example} illustrates the difference between standard in-context prompting and instruction-induction prompting.

To elicit models to generate instructions, we consider prompts that would elicit humans to do so. We design a meta-prompt presenting instruction induction as a challenge puzzle and verify its clarity in a human study (\S\ref{sec:verification}).
The prompt is presented in Figure~\ref{fig:induction_example} (right side, in pink).\footnote{We found this prompt informative for both humans and models in preliminary experiments.}

While prior work already shows that large LMs are often able to infer a latent task from a given set of demonstrations, this has been largely based on their ability to \textit{execute} the task on a held-out example.
Instruction induction requires that the model \textit{describe} the underlying task in natural language.

%% file: 03_data.tex
\section{Data}
\label{sec:data}

We evaluate on 24 tasks, listed in Table \ref{tab:tasks}.
We select these tasks as they vary in difficulty and represent different aspects of language understanding, ranging from surface-level spelling to sentence similarity and causality detection.\footnote{See Appendix~\ref{sec:data_construction} for the full details of each task.}
We review the dataset's format, the annotation and verification processes we conducted to ensure that the tasks are viable, and finally discuss a theoretical limitation of this setup.

\input{03a_data_tab}

\subsection{Format}

In every task, each single \textit{demonstration} $(x_k, y_k)$ is formatted as follows:
\begin{center}
Input: $x_k$ ~~\\
Output: $y_k$
\end{center}
For instance, one demonstration in the pluralization task is ``Input: cat'' followed by ``Output: cats'' in a new line.
We split each task's demonstrations into two sets: an \textit{induce} set, which we use for generating instructions, and an \textit{execute} set, which is held out for the execution accuracy evaluation metric (see \S\ref{sec:execution}).
Each \textit{instruction induction example} is composed of 5 demonstrations sampled randomly without replacement from the induce set, concatenated with new-line separators; we create 100 examples for each task.
When generating instructions, each example is placed inside the instruction induction prompt, and fed to the model (Figure~\ref{fig:induction_example}, right).

\subsection{Annotating Reference Instructions}
\label{sec:annotations}

We collect 10 gold-reference human-annotated instructions via college-graduate English-speaking annotators.
For each task, we provide the annotators with the exact same input we intend to provide a model: 5 input-output demonstrations wrapped by the instruction-induction prompt (Figure \ref{fig:induction_example}).
We manually verify each annotation and discard ones that do not correctly describe the task. 
We refer to this set of annotations as the \textit{gold} annotations, and use them for reference-based evaluation (see \S\ref{sec:eval}).


\subsection{Verification}
\label{sec:verification}

Prior to the instruction induction experiments, we conduct two tests to ensure that either models or humans can infer the underlying task given 5 demonstrations.
We first verify that models can indeed execute our tasks given 5 demonstrations using in-context learning.
Secondly, we conduct a human study to confirm that 5 demonstrations are enough for humans to describe the latent tasks.

\paragraph{In-Context Learning}
We prompt models with 5 input-output demonstrations and concatenate an additional test input $x_{k+1}$,
and verify that the models are able to correctly predict $y_{k+1}$ (Figure~\ref{fig:induction_example}, left).
For each task, we repeat this experiment 100 times, each with a different set of demonstrations and test inputs.
We do not provide the model with any instruction beyond the ``Input: $x_k$ Output: $y_k$'' format. We evaluate each task using its predefined evaluation metric.\footnote{All metrics are variants of simple string matching, with some task-specific heuristics, for example, to allow for multiple correct answers. See Appendix~\ref{sec:data_construction} for exact details.}
The in-context results for GPT-3 \citep{gpt3} and InstructGPT \citep{instruct-gpt} (see model details in \S\ref{sec:results}) are reported in Table \ref{tab:verification_tab} in Appendix~\ref{sec:data_verification}, which shows that in-context learning can reach 80\% accuracy and above on most tasks.

\paragraph{Human Study}
To assess the human ability to induce instructions, we collect human-written instructions, using annotators that \textit{did not} participate in the gold references collection. As in the gold-reference annotation process, we provide annotators with the same input we intend to provide to models. We refer to this set of annotations as the \textit{control} annotations. We then manually count, for each task, the number of annotators that provided a correct instruction, and report the correct instructions percentage in Table \ref{tab:verification_tab} (Appendix~\ref{sec:data_verification}). 
In all but one task (\textit{Larger Animal}), at least 4 out of 5 annotators were able to produce correct task descriptions.

We also use the control group's annotations to establish a human baseline for automatic evaluation metrics.
For reference-based evaluation (\S\ref{sec:bertscore}), we treat the control annotations as generated instructions and compare them against the gold annotations, while for execution accuracy (\S\ref{sec:execution}), we use the control annotations to measure human performance, and the gold references as a ceiling metric.

\subsection{Ambiguity}
\label{sec:ambiguity}

A theoretical challenge in inducing instructions is ambiguity.
For example, when given the single demonstration ``Input: The coffee is too hot. Output: The, too, hot'', one could infer that the underlying task is either ``write all the words containing the letter T'' or ``write all the three-lettered words'', both valid interpretations.
Ambiguity might confuse models tasked with instruction induction while also making evaluation less reliable.
In practice, providing 5 demonstrations typically resolves the ambiguity in our set of tasks.
As evident from the data verification process, our tasks can typically be inferred by models and/or humans.

Inducing more complex task descriptions, such as predicting detailed annotation guidelines, may pose a greater challenge in terms of ambiguity. We hypothesize that providing more than 5 demonstrations could mitigate some of that challenge, and leave further exploration of this avenue to future work.

%% file: 03a_data_tab.tex
\begin{table}[p]
\small
\centering
\begin{tabular}{@{}p{0.12\textwidth}@{}p{0.175\textwidth}@{}p{0.375\textwidth}p{0.300\textwidth}@{}}
\toprule
\textbf{Category}                           & \textbf{Task}           & \textbf{Instruction}                                             & \textbf{Demonstration} \\
\midrule
\textit{Spelling} & First Letter & Extract the first letter of the input word. & cat $\rightarrow$  c \\
\cmidrule{2-4}
 & Second Letter & Extract the second letter of the input word. & cat $\rightarrow$  a \\
\cmidrule{2-4}
 & List Letters & Break the input word into letters, separated by spaces. & cat $\rightarrow$  c a t\\
\cmidrule{2-4}
& Starting With & Extract the words starting with a given letter from the input sentence. & The man whose car I hit last week sued me. [m] $\rightarrow$  man, me \\
\midrule
\textit{Morpho-}

\textit{syntax} & Pluralization  & Convert the input word to its plural form.                   & cat $\rightarrow$  cats \\
\cmidrule{2-4} 
                                   & Passivization  & Write the input sentence in passive form.             &
The artist introduced the scientist. $\rightarrow$  The scientist was introduced by the artist. \\
\midrule
\textit{Syntax}                             & Negation       & Negate the input sentence.   & Time is finite $\rightarrow$  Time is not finite. \\
\midrule
\textit{Lexical} 

\textit{Semantics} & Antonyms       & Write a word that means the opposite of the input word. & won $\rightarrow$ lost \\
\cmidrule{2-4}
                                   & Synonyms       & Write a word with a similar meaning to the input word.  & alleged $\rightarrow$  supposed\\
\cmidrule{2-4}
                                   & Membership    & Write all the animals that appear in the given list.    & cat, helicopter, cook, whale, frog, lion $\rightarrow$  frog, cat, lion, whale \\
\midrule
\textit{Phonetics}                          & Rhymes         & Write a word that rhymes with the input word.           & sing $\rightarrow$  ring \\
\midrule
\textit{Knowledge}                    & Larger Animal  & Write the larger of the two given animals.              & koala, snail $\rightarrow$  koala\\
\midrule
\textit{Semantics} & Cause Selection & Find which of the two given cause and effect sentences is the cause. & Sentence 1: The soda went flat. Sentence 2: The bottle was left open. $\rightarrow$  The bottle was left open.\\
\cmidrule{2-4}
& Common

Concept & Find a common characteristic for the given objects. & guitars, pendulums, neutrinos $\rightarrow$  involve oscillations.\\
\midrule
\textit{Style} & Formality & Rephrase the sentence in formal language. & Please call once you get there $\rightarrow$  Please call upon your arrival.\\
\midrule
\textit{Numerical}         & Sum            & Sum the two given numbers.   & 22 10 $\rightarrow$  32 \\
\cmidrule{2-4}
                                   & Difference     & Subtract the second number from the first.  & 32 22 $\rightarrow$  10 \\
\cmidrule{2-4}
                                   & Number to Word & Write the number in English words.  & 26 $\rightarrow$  twenty-six \\
\midrule
\textit{Multi-}

\textit{lingual} & Translation    & Translate the word into German / Spanish / French.    & game $\rightarrow$  juego\\
\midrule
\textit{GLUE} & Sentiment 

Analysis & Determine whether a movie review is positive or negative. & The film is small in scope, yet perfectly formed. $\rightarrow$  positive \\
\cmidrule{2-4}
& Sentence 

Similarity & Rate the semantic similarity of two input sentences on a scale of 0 - definitely not to 5 - perfectly. & Sentence 1: A man is smoking. Sentence 2: A man is skating. $\rightarrow$  0 - definitely not \\
\cmidrule{2-4}
& Word in Context & Determine whether an input word has the same meaning in the two input sentences. & Sentence 1: Approach a task. Sentence 2: To approach the city. Word: approach  $\rightarrow$  not the same \\
\bottomrule
\\
\end{tabular}
\caption{The tasks in our instruction-induction benchmark. For each task, we show a corresponding instruction and demonstration, with $\rightarrow$ separating the input from the output.}
\label{tab:tasks}
\end{table}

%% file: 04_evaluation.tex
\section{Evaluating Generated Instructions}
\label{sec:eval}
As a standard text generation metric, we report BERTScore \citep{bertscore}. However, the instruction induction challenge has a unique property, which does not usually hold for other text generation tasks: the instructions are \textit{executable}. Their correctness can therefore be measured directly by utilizing them as prompts.


\subsection{Reference-Based Evaluation}
\label{sec:bertscore}

We use BERTScore \citep{bertscore} to compare the model-generated instructions against the collected gold annotations.
As mentioned in \S\ref{sec:annotations}, we use only the correct, verified annotations as references. 
We take the maximal BERTScore-F1 over all gold-reference annotations to account for natural variations in instruction formulation.\footnote{We use BERTScore version 0.3.11 with the DeBERTa-xl-MNLI model \citep{he2021deberta, nangia-etal-2017-repeval}.}
We also establish a human baseline for each task using the \textit{control} annotations, which were collected from a separate control group of annotators (\S\ref{sec:data_verification}), which we compare against the \textit{gold} annotations in exactly the same way as model-generated instructions.

\subsection{Execution Accuracy}
\label{sec:execution}

We introduce \textit{execution accuracy}, a new metric unique to the instruction induction task.
To measure the execution accuracy of a predicted instruction $I$ (e.g., ``Write the plural form of the given word.'') for a task $T$ (pluralization), we prompt a model with $I$ and an input $x$ (``cat'').
We then test, given $I$ and $x$, whether the model can correctly predict $y$, the output of performing $T$ on the input $x$ (\textit{cats}).

To obtain meaningful results, we measure execution accuracy on the 100 held-out \textit{execute} examples for each task.
The execution accuracy of an instruction $I$ is therefore computed by taking the average over $Score_{T}(I(x_n),y_n)$ for all $x_n$ in the \textit{execute} set, where $Score_{T}$ denotes the task's corresponding metric (see Appendix~\ref{sec:data_construction}), and $I(x_n)$ is the result of prompting a predefined language model with the instruction $I$ and the input $x_n$.
As recent models are trained to follow instructions \citep{sanh2022multitask,wei2022finetuned,instruct-gpt}, and due to the relative clarity of our tasks, we expect correct instructions to yield high execution accuracy when using a sufficiently powerful execution model.

%% file: 05_results.tex
\begin{figure}[h!]
\centering
 \includegraphics[width=1.0\textwidth]{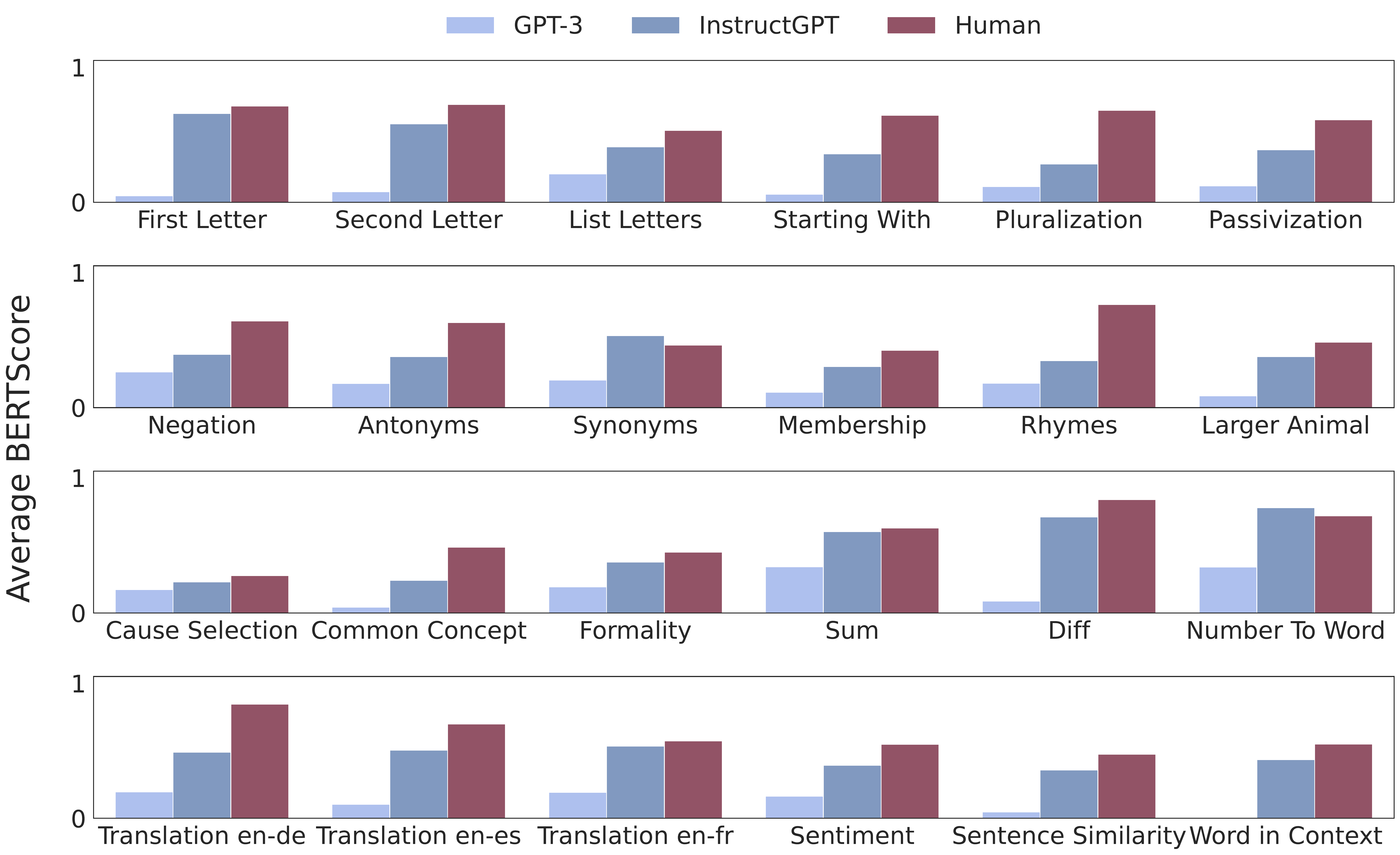}
 \caption{Average BERTScores of model-generated instructions for each task, compared to the performance of the control group's manually-authored instructions. The BERTScore for each instruction is computed using the human \textit{gold} annotations as references.}
 \label{fig:bertscore}
\end{figure}

\begin{figure}[h!]
\centering
 \includegraphics[width=1.0\textwidth]{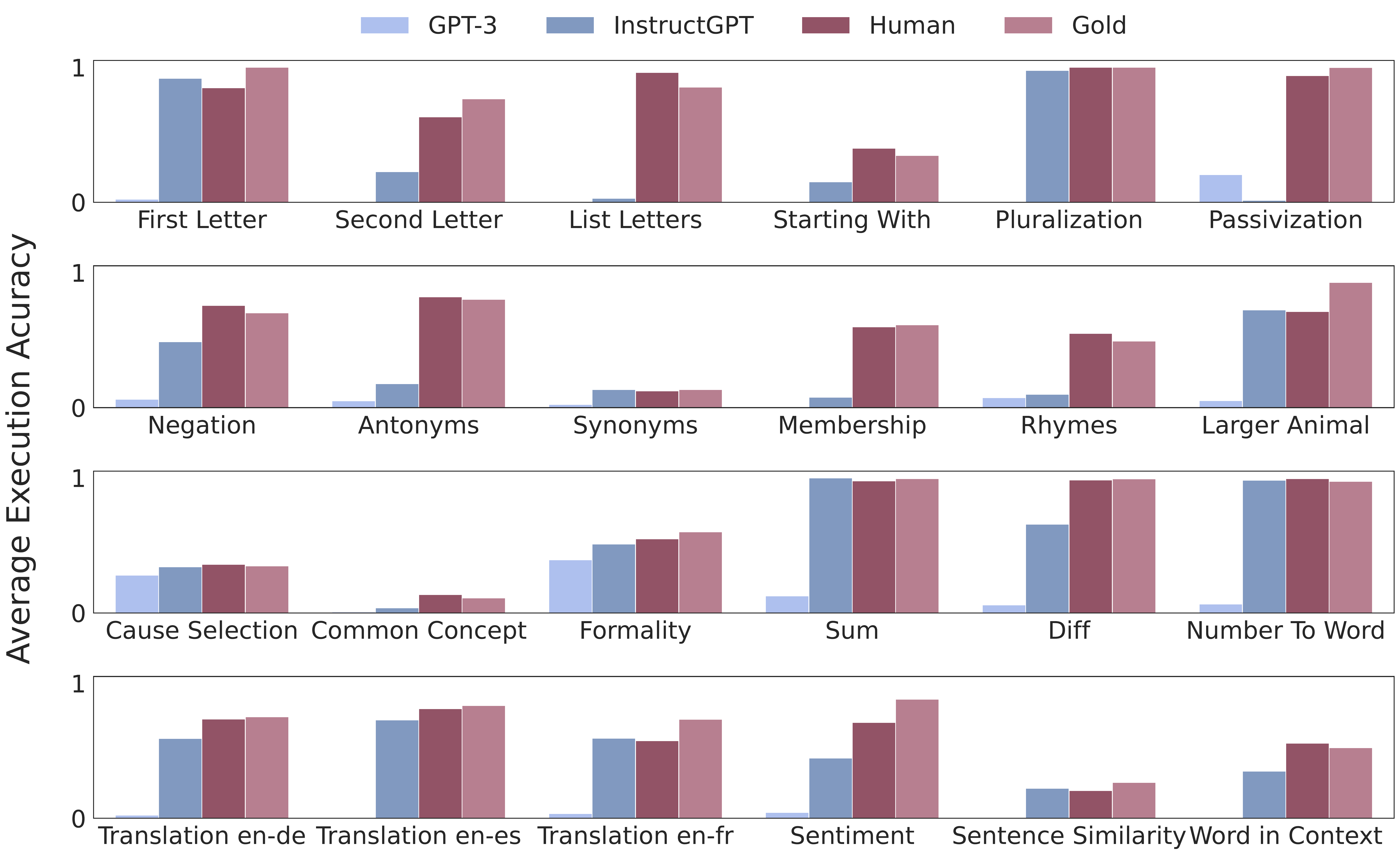}
 \caption{Average execution accuracy of model-generated instructions for each task, compared to the execution accuracy measured for human-written instructions. The \textit{Human} baseline is measured by taking the control group's annotations, while the \textit{Gold} ceiling metric is based on the separately-annotated and verified gold annotations.}
 \label{fig:execution_acc}
\end{figure}

\section{Results}
\label{sec:results}

\paragraph{Baseline Models}
We experiment with eight versions of GPT-3 \citep{gpt3}, a Transformer decoder language model.
First, we experiment with the most current version available in the OpenAI API, for each of the four available model sizes.
Though not stated explicitly in the API, we assume these models are those reported by \citet{instruct-gpt}, and we therefore refer to them as \textit{Instruct} models.\footnote{Concretely, we use: text-davinci-002, text-curie-001, text-babbage-001, text-ada-001.}
We also experiment with the four originally published GPT-3 versions.\footnote{davinci, curie, babbage, ada.} 
By default, we refer to the largest Instruct model as \textit{InstructGPT}, and the original 175B-parameter model as \textit{GPT-3}.
All model generations were produced using the greedy decoding algorithm.

\begin{table}[t]
\small
\centering
\begin{tabular}{@{}lrr@{}}
\toprule
\textbf{Model} & \textbf{BERTScore} & \textbf{Execution} \\
\midrule
\textit{GPT-3} & & \\
~~~~Ada & -7.7 & 4.0 \\
~~~~Babbage & 4.1 & 3.2 \\
~~~~Curie & 13.9 & 7.9 \\
~~~~DaVinci & 14.6 & 6.5 \\
\midrule
\textit{InstructGPT} & & \\
~~~~Ada & 5.9 & 4.4 \\
~~~~Babbage & -0.5 & 3.8 \\
~~~~Curie & 10.7 & 8.8 \\
~~~~DaVinci & 44.4 & 43.6 \\
\midrule
\textit{Human (Control)}  & 60.0 & 66.4 \\
\bottomrule
\\
\end{tabular}
\caption{Average BERTScore and execution accuracy across tasks. BERTScore is measured against the gold references. The execution accuracy for all generated instructions is measured using InstructGPT as the execution model. Human performance is measured using the human control group's instructions.}
\label{tab:averages}
\end{table}
\label{sec:results_bertscore}

\subsection{Comparing to Gold Annotations}

Figure~\ref{fig:bertscore} presents the average BERTScore per task (see \S\ref{sec:bertscore}).
Results show that the InstructGPT model has, to some extent, the ability to induce instructions from a few demonstrations; in 13 out of 24 tasks it achieves at least 75\% of human performance.
GPT-3, on the other hand, is quite far from human performance across the board.

Table~\ref{tab:averages} shows the average scores across all tasks. We observe the same trend; while InstructGPT's BERTScore is 15.6 points lower than human performance, the gap between GPT-3 and humans is 45.4 points.
Moreover, we observe that smaller models -- even those fine-tuned to follow instructions -- do not exhibit any instruction-induction abilities. Scores are slightly higher for larger models of the same family (except for the InstructGPT-Babbage outlier), but are overall low.
Excluding the largest models, there does not appear to be a significant advantage for Instruct models over the originals when controlling for model size.

\subsection{Execution Accuracy}\label{sec:results_execution}

We compute the execution accuracy as detailed in \S\ref{sec:execution}, and report the average over 100 generated instructions for each task.
As an execution model, we use the largest InstructGPT model.
We also use this model to induce instructions, and while using it as an execution model might bias results towards its own generations, preliminary experiments show that no other model is as good at following instructions as InstructGPT.
As a point of reference, we apply the execution accuracy evaluation protocol to human-written instructions.
First, to compare models with human performance, we measure the execution accuracy of the \textit{control} annotation set.
Second, to account for limitations in the execution model, we measure execution accuracy of the correct (manually verified) \textit{gold} annotations, which acts as an approximated ceiling metric.

Figure~\ref{fig:execution_acc} presents the execution accuracy per task.
In 12 out of 24 tasks, InstructGPT achieves at least 75\% of the execution accuracy measured for the human-written instructions.
GPT-3 shows much weaker execution accuracy, scoring less than 10\% on 20 of the 24 tasks.
In fact, only in the cases of formality, passivization, and cause selection does it approach human performance, and that is largely an artifact of a more lenient evaluation metric in the case of formality and cause selection, or due to the execution model being right for the wrong reasons in the case of passivization (see \S\ref{sec:analysis}).
In some tasks, the control annotations are of high quality and reach a higher score than the verified gold annotations, likely due to variance of the execution model in such cases.

Table~\ref{tab:averages} shows the same trends. On average, InstructGPT achieves 65.7\% of human performance, while GPT-3 reaches only 9.8\% of human performance. When considering different model families or sizes, we do not see any substantial improvements when increasing model size or adding instruction tuning, with the exception of the largest InstructGPT model. The ability to generate instructions seems to only emerge when a model is both large enough and aligned to follow instructions.
Overall, even the best-performing model still does not reach human performance, leaving room for future improvement.



%% file: 06_analysis.tex
\section{Analysis}
\label{sec:analysis}

\begin{table}[t]
    \small
    \centering
    \begin{tabular}{@{}p{0.18\textwidth}p{0.35\textwidth}p{0.41\textwidth}@{}}
    \toprule
    \textbf{Task} & \textbf{GPT-3} & \textbf{InstructGPT} \\
    \midrule
    First letter & The friend's output was: & Write the first letter of each word. \\
    \midrule
    Sentence Similarity & The friend wrote the following output: & For each input, rate the similarity of the two sentences on a scale of 0 to 5, with 5 being a perfect match. \\
    \midrule
    Pluralization & The friend's output was: & Add `s' to the end of each word. \\
    \midrule
    Passivization & The friend wrote the following output: & Reverse the order of the subject and the object in the sentence. \\
    \midrule
    Antonyms & The friend's output was: & Reverse the input. \\
    \bottomrule
    \\
    \end{tabular}
    \caption{Examples of the instructions generated by GPT-3 and InstructGPT for five of our tasks.}
    \label{tab:analysis}
\end{table}

To gain further insight into the successes and failures of instruction induction prompting, we manually analyze the model-generated instructions of 5 tasks.
Table \ref{tab:analysis} shows the most common predictions of GPT-3 and InstructGPT for each of these tasks.

InstructGPT obtains high, or close to human execution accuracy scores for three of these tasks (\textit{First Letter}, \textit{Sentence Similarity}, \textit{Pluralization}).
Indeed, the instructions for both \textit{First Letter} and \textit{Sentence Similarity} accurately describe the task.
However, the instruction generated for \textit{Pluralization} is not entirely precise, since it dismisses other forms of pluralization such as -es, -ies, and irregulars.
Although the instruction only asks to add an ``s'', the execution model often ignores the specifics and produces the correct plural form; in one case, the input word was ``life'' and the output was ``lives''.
While this particular instruction accounts for 24\% of the induced instructions in the pluralization task, some predictions do explicitly mention pluralization, though not always accurately, e.g., ``Add -s to the end of each word to make it plural''.

For some tasks, InstructGPT fails to produce accurate instructions, even if it is able to solve via in-context learning (see Table~\ref{tab:verification_tab}).
In \textit{Passivization}, 98\% of the predicted instructions were to simply ``reverse the order of the subject and object'', while ignoring additional surface-form manipulations needed to convert the given sentence into passive form; e.g., for the input ``The authors supported the scientist'', following the instructions produces the output ``The scientist supported the authors'', while the correct passive form is ``The scientist was supported by the authors''.
Surprisingly, the instructions generated by GPT-3 obtained higher execution accuracy than the InstructGPT, even though they were entirely unrelated.
In 24\% of the cases, GPT-3 predicted ``The friend wrote the following output:'' -- an instruction that apparently prompts the execution model to often rephrase the input in passive form.
Lastly, in \textit{Antonyms}, 60\% of InstructGPT's predictions were ``Reverse the input'', and another 11\% were ``Reverse the word''.
While one could imagine an interpretation of these instructions that reflects the task (reversing the \textit{meaning} of the word), the execution model interprets them literally, and reverses the input words' letters.

Overall, GPT-3 did not exhibit any instruction induction abilities, although it did often phrase outputs in imperative language.
One relatively common prediction was the generic instruction ``Write an output for every input''.
Because these empty instructions are in the right format, they tend to have some overlap with the reference instructions, which inflates their BERTScore.
Execution accuracy, on the other hand, is robust to this phenomenon, and typically assigns GPT-3's outputs very low scores.

%% file: 07_related_work.tex
\section{Related Work}

\paragraph{In-Context Learning}
\citet{gpt3} suggest that models can learn a task by conditioning on few input-output demonstration pairs, without any fine-tuning or gradient updates. This paradigm, known as \textit{in-context learning} or \textit{prompt-based learning} \citep{prompt-based}, has been the focus of many research efforts lately:
\citet{glam} suggest methods for more efficient in-context learning,
\citet{zhao-2021-calibrate} study methods for improving the stability and accuracy of prompt-based models, \citet{chen-meta-incontext} and \citet{min2022metaicl} conduct meta-training with an in-context learning objective,
while other work studies the effect of the provided prompts \citep{reynolds-2021-prompt-programming, webson-pavlick-2021, min2022rethinking}, or suggests prompt reframing techniques \citep{reframing-prompts} and prompt retrieval methods \citep{prompt-retrieval}.
To the best of our knowledge, all previous work study in-context learning through the lens of \textit{executing} a latent task, while we focus on the ability to explicitly \textit{describe} it.

\paragraph{The Instruction Paradigm}
\citet{turking-test} propose to learn new tasks from natural language instructions.
\citet{naturalinstructionsv1} and \citet{naturalinstructionsv2} collect crowdsourcing instructions used to create NLP datasets into a benchmark for measuring the ability to solve tasks by reading instructions.
Recent work shows that fine-tuning on task instructions (\textit{instruction tuning}) improves the zero-shot learning abilities of LMs \citep{sanh2022multitask,wei2022finetuned,instruct-gpt}. This work focuses on models' ability to \textit{generate} instructions, rather than their ability to \textit{execute} instructions written by humans.

\paragraph{Intermediate Reasoning Steps}
\citet{nye2022show} show that LMs can perform complex computations by writing intermediate steps on a ``scratchpad''.
In \textit{chain of thought prompting} \citep{wei2022chain}, input-output demonstrations are enriched with sentences elaborating intermediate task reasoning steps, improving the performance of LMs on tasks requiring reasoning skills.
Subsequent work further improves the performance on such tasks using a \textit{self-consistency} ensemble \citep{self-consistency-cot}, which samples a set of diverse chain-of-thought reasoning paths, taking the majority vote over all generated answers.
\citet{star-cot} utilize a small set of examples labeled with chain-of-thought rationales and a large set of unlabeled data to iteratively bootstrap automatic rationale generation, thus creating a large dataset labeled with such rationales to enable fine-tuning.
In contrast, we study the ability of LMs to generate a description of the task, rather than generating intermediate reasoning steps as a means of executing complex tasks.






%% file: 08_discussion.tex
\section{Discussion}

This work demonstrates that large LMs can not only infer new tasks based on a handful of demonstrations, but also describe them in natural language.
We provide evidence of this ability on a diverse set of language tasks, and show that while instruction induction abilities are limited to a single state-of-the-art model, this model does indeed approach human performance on about half the tasks.

It is not unreasonable to assume that models in the near future will be even better at processing human-generated instructions, and it is therefore interesting to discuss the potential applications of instruction induction.
In particular, we envision a use case in which instruction induction serves as a machine learning approach; instead of converting a dataset into a set of continuous parameters, we could produce a natural language instruction that best describes the data. Grounding the model in concise natural language has the advantage of interpretability, and has the potential to solve fundamental issues pertaining to spurious correlations. While it is still too early to determine whether this approach is viable, we view it as an intriguing direction for future research.



%% file: 09_appendix.tex


\section{Dataset Details}
\label{sec:data_construction}

This appendix details each task's dataset (\S\ref{sec:data_construction_details}).
Some datasets rely on a set of common English nouns (CEN), described at \S\ref{sec:common_nouns}.

\subsection{Tasks}
\label{sec:data_construction_details}

We elaborate on each task's data source, preprocessing protocol, and evaluation metric used in the in-context learning and execution accuracy experiments.
As mentioned in \S\ref{sec:data}, each task has \textit{induce} and \textit{execute} sets; unless stated otherwise, we sample 100 examples as the execute set for each task.
When evaluating outputs, the generated text is first normalized; we take only the first generated sentence and  lowercase it.
We apply exact string match as the evaluation metric where applicable, elaborating only where alternative metrics are used.

\paragraph{First Letter}
In each demonstration, $x_k$ is a noun, and $y_k$ is the first letter of that noun. We construct the demonstrations by extracting the first letter of each word in CEN.

\paragraph{Second Letter}
Identical to the \textit{First Letter} task, only here $y_k$ is the second letter of $x_k$.

\paragraph{List Letters}
$x_k$ is a noun from CEN, and $y_k$ is a list of $x_k$'s letters, separated by spaces.

\paragraph{Starting With}
$x_k$ contains a sentence and a letter in brackets, and $y_k$ lists the words in $x_k$ that start with the given letter. We avoid cases in which $y_k$ is empty, i.e., there is always at least one word in the input sentence starting with the given letter.
Sentences are taken from the CoLA dataset \citep{warstadt2018neural}.
For the induce set, we create all (sentence, letter) pairs using CoLA’s train set, and then sample 3,000 pairs.
For the \textit{execute} set, we create all (sentence, letter) pairs from CoLA’s in-domain and out-of-domain dev sets, and then sample 50 in-domain and 50 out-of-domain examples.
We evaluate using exact \textit{set} match, by treating the output (and $y_k$) as a set of strings.

\paragraph{Pluralization}
Given a singular noun $x_k$, produce the plural form $y_k$.
We take noun inputs from the CEN set, filtering out mass nouns using a predefined list.\footnote{\url{https://gist.github.com/sudodoki/b5408fa4ba752cc22597250fc58a5970}}
To create the plural forms, we apply an automatic pluralization engine\footnote{\url{https://pypi.org/project/inflect/}} and exclude nouns for which the engine's output did not appear at least 50 times in the Wikitext-103 corpus.
This results in 2,043 singular-plural noun pairs.

\paragraph{Passivization}
Given a simple active sentence $x_k$, rephrase the sentence in passive voice $y_k$.
We use the 1,000 HANS \citep{mccoy-etal-2019-right} evaluation set active-passive entailed sentence pairs.

\paragraph{Negation}
$y_k$ is the negation of the input sentence $x_k$.
We use the negated LAMA dataset \citep{petroni-etal-2019-language,kassner-schutze-2020-negated}, taking the 304 negated SQuAD \citep{rajpurkar-etal-2016-squad} sentences, 300 ConceptNet \citep{speer-havasi-2012-representing} sentences, 200 T-REx \citep{elsahar-etal-2018-rex} sentences and 200 Google-RE\footnote{\url{https://code.google.com/archive/p/relation-extraction-corpus/}} sentences. For ConceptNet and T-REx, we manually select these sentences to ensure their quality. For Google-RE, we automatically sample 100 sentences from the \textit{place of birth} relation, and 100 from the \textit{place of death} relation.

\paragraph{Antonyms}
$y_k$ is the antonym of the input word $x_k$.
We use the antonym pairs from oLMpics \citep{talmor-etal-2020-olmpics}, which were extracted from ConceptNet \citep{speer-havasi-2012-representing} and WordNet \citep{fellbaum1998wordnet}.
For uniformity, we verify that all pairs are indeed antonyms according to WordNet.

\paragraph{Synonyms}
$x_k$ is a word and $y_k$ is its synonym.
As in the antonyms task, we use the synonym pairs of \citet{talmor-etal-2020-olmpics}.
Since there can be multiple synonyms for each input word, the task's in-context and execution accuracy are evaluated by testing whether the gold answer (a single word) is contained in the predicted answer (which may be a list of words).

\paragraph{Membership}
$x_k$ is a list of words, where some of the words represent animals, and $y_k$ lists the animals from $x_k$.
To construct the task's data, we first select 6 word categories: animals, clothing, colors, food, vehicles, and professions.
We then take 10-50 words from each category, using only words that are categorized at the A1 or A2 levels according to the Common European Framework of Reference for Languages (CEFR).\footnote{https://languageresearch.cambridge.org/american-english}
Using these words, we create random lists containing between 5 to 7 words, where 3 or 4 are animals and the rest belong to one of the other 5 categories. The induce split is constructed by sampling 3,000 such combinations, using 80\% of each category's words. The execute split is constructed by sampling 100 such combinations, using the remaining 20\% of each category's words.
The task's in-context and execution accuracy are evaluated using an exact \textit{set} match, by treating the output (and $y_k$) as a set of strings.

\paragraph{Rhymes}
$y_k$ is a rhyme of the input word $x_k$.
The data was constructed by taking words categorized at the A1, A2, or B1 levels according to CEFR.
We then use CMU's pronouncing dictionary\footnote{\url{https://github.com/cmusphinx/cmudict}} to find rhyming groups for these words.
The execute split is constructed by sampling 30 rhyming groups, each containing two or more words, and sampling 100 unique words. The induce split is constructed using the rest of the rhyming groups.
We evaluate this task by checking whether the predicted word is contained in the rhyming group of $x_k$.

\paragraph{Larger Animal}
$x_k$ is two animals, and $y_k$ is the (physically) larger one.
We use the object comparison data from oLMpics \citep{talmor-etal-2020-olmpics}, taking the train split, which only contains animals.
We construct the induce set using a sample of 80\% of the animals and the execute set by sampling 100 pairs out of the remaining 20\% animals.

\paragraph{Cause Selection}
$x_k$ contains two sentences describing related events, where one event caused the other; $y_k$ contains the cause sentence.
As data source, we use the 50 examples from the BIG-bench \citep{bigbench} \textit{Cause and Effect} task, randomly splitting them to equally-sized induce and execute sets.
In each of the induce demonstrations, we randomly sample the position of the cause sentence (either the first or the second sentence in $x_k$).
For examples in the execute set, we take both options for each cause and effect pair, doubling the data.

\paragraph{Common Concept}
$x_k$ contains a few entities that share a non-trivial common underlying concept, while $y_k$ describes that common concept.
We use the 32 examples from \textit{Novel Concepts} in BIG-bench \citep{bigbench}, using half for induce and half for execute.
As the BIG-bench answers usually contain clear ``task markers'' (e.g., answers that start with ``They all have...'', indicating that the task was to find a common concept), we remove them from our demonstrations.
The task's in-context and execution accuracy are evaluated using unigram overlap (F1).

\paragraph{Formality}
$x_k$ is a sentence in informal English, and $y_k$ is its paraphrase in more formal language.
We write 30 sentence pairs ourselves, following existing guidelines for converting informal sentences into formal ones.\footnote{\url{https://www.niu.edu/writingtutorial/style/formal-and-informal-style.shtml}, \url{https://www.uts.edu.au/current-students/support/helps/self-help-resources/grammar/formal-and-informal-language}}
The task's in-context and execution accuracy are evaluated using unigram overlap (F1).

\paragraph{Sum}
$x_k$ contains two numbers separated by a space, and $y_k$ is their sum.
For each number in the range $[0,99]$, we enumerate over all pairs.

\paragraph{Difference}
$x_k$ contains two numbers separated by a space, and $y_k$ is the difference between them.
We use all number pairs such that both input numbers are in the range $[0,198]$, and always subtract the smaller number from the bigger number.

\paragraph{Number to Word}
$x_k$ is a number written in digits (e.g., 28), and $y_k$ is the same number written in words (e.g, twenty-eight).
We use all numbers in range [0,9999].

\paragraph{Translation}
$x_k$ is an English word and $y_k$ is its translation to some target language -- either German, Spanish, or French. We use CEN as input words, and obtain their translations via Wiktionary.\footnote{\url{https://github.com/open-dsl-dict/wiktionary-dict}}
For evaluation, we check whether the predicted answer is contained in the set of the possible gold answers.

\paragraph{Sentiment Analysis}
$x_k$ is a movie review and $y_k$ is a binary label, either ``positive'' or ``negative'', marking the review's sentiment.
We use the Stanford Sentiment Treebank dataset \citep{socher-etal-2013-recursive} from GLUE \citep{wang-etal-2018-glue}, taking the train split as our induce set and the dev split as the execute set.
We consider only full sentences, discarding sentence constituents and sentences containing more than 10 words.
This leaves us with an induce set of 1,167 examples. To create label-balanced instruction induction examples, we sample each sequence of 5 demonstrations such that there are at least 2 demonstrations for each label.

\paragraph{Sentence Similarity}
$x_k$ contains two sentences, and $y_k$ reflects the semantic similarity of the two input sentences.
The similarity is measured on a scale of 0 to 5, and the labels contain an additional short textual description of the numerical label, e.g., ``5 - perfectly''.
We use the Semantic Textual Similarity Benchmark dataset \citep{cer-etal-2017-semeval} from GLUE, rounding the similarity scores and taking the train split as the induce set and the dev split as the execute set.
We discard examples in which at least one of the sentences contains more than 10 words, which leaves us with an induce set of 3,716 examples.
In each instruction induction example, we sample at least one pair with a score of 0 and one with a score of 5, so that models will be exposed to the minimal and maximal scores when generating an instruction.
We evaluate whether the predicted answer matches one of three valid outputs for each label: the numerical label (``5''), the verbal label (``perfectly''), or the combined label (``5 - perfectly'').

\paragraph{Word in Context}
$x_k$ contains a target word and two contexts (sentences) for that word, and $y_k$ is a binary label reflecting whether the word has the same meaning in both contexts.
We use the Word in Context dataset \citep{pilehvar-camacho-collados-2019-wic} from SuperGLUE \citep{superglue}, taking the train split as the induce set and the dev split as the execute set.
We discard examples in which at least one of the sentences contains more than 10 words, which leaves us with an induce set of 4,084 examples.
To create label-balanced instruction induction examples, we sample each sequence of 5 demonstrations such that there are at least 2 demonstrations for each label.
We evaluate whether the predicted label matches one of several possible outputs: ``same'', ``yes'', or ``true'' for an identical meaning, and ``not the same'', ``no'', or ``false'' for a different meaning.

\subsection{Common English Nouns}
\label{sec:common_nouns}

We create a dataset of common English nouns (CEN) by filtering high-frequency nouns from the Wikitext-103 corpus \citep{merity2016pointer}.
We first create a vocabulary of the 10,000 most frequent words in the corpus, from which we will later select the nouns.
We then process the corpus with SpaCy's part-of-speech tagger and lemmatizer,\footnote{\url{https://spacy.io/}}
and retain only nouns that appear in their singular form by verifying that their part-of-speech tag is ``NN'' and testing whether the word's lemma is identical to the word itself.
We additionally filter nouns that have less than 3 letters. Overall, this leaves us with a set of 3,406 nouns.

\section{Data Verification}
\label{sec:data_verification}

Table~\ref{tab:verification_tab} shows the results for the data verification experiments (\S\ref{sec:verification}). As evident by these results, most of our tasks can be inferred in-context by models. Moreover, all tasks but one can be accurately described by at least 4 out 5 human annotators.

\input{03b_in_context_tab}

%% file: 03b_in_context_tab.tex

\begin{table}[t]
\small
\centering
\begin{tabular}{@{}lrrr@{}}\toprule
\textbf{Task} & \multicolumn{2}{c}{\textbf{In-Context Learning}} & \textbf{Human Study}
\\\cmidrule(lr){2-3} & GPT-3  & InstructGPT & \\
\midrule
First Letter & \textbf{97} & \textbf{98} & \textbf{100} \\
Second Letter & 25 & 34 & \textbf{100} \\
List Letters & \textbf{98} & \textbf{100} & \textbf{100} \\
Starting With & 33 & 46 & \textbf{80} \\
\midrule
Pluralization & \textbf{95} & \textbf{99} & \textbf{100} \\
Passivization & \textbf{100}  & \textbf{100}  & \textbf{80}  \\
\midrule
Negation & \textbf{94} & \textbf{93} & \textbf{100}  \\
\midrule
Antonyms & \textbf{84} & \textbf{83} & \textbf{100}  \\
Synonyms & 9 & 12 & \textbf{80} \\
Membership & 13 & 36 & \textbf{100} \\
\midrule
Rhymes & 46 & 39 & \textbf{100} \\
\midrule
Larger Animal & 58 & \textbf{82} & 40 \\
\midrule
Cause Selection & 47 & \textbf{82} & \textbf{100} \\
Common Concept & 23 & 15 & \textbf{100} \\
\midrule
Formality & 54 & 56 & \textbf{80} \\
\midrule
Sum & \textbf{87} & \textbf{100} & \textbf{100} \\
Diff & 69 & \textbf{95} & \textbf{100} \\
Number To Word & \textbf{85} & \textbf{100} & \textbf{100} \\
\midrule
Translation en-de & \textbf{80} & \textbf{85} & \textbf{100} \\
Translation en-es & \textbf{91} & \textbf{88} & \textbf{100} \\
Translation en-fr & \textbf{80} & \textbf{84} & \textbf{80} \\
\midrule
Sentiment & \textbf{95} & \textbf{99} & \textbf{100} \\
Sentence Similarity & 3 & 15 & \textbf{80} \\
Word in Context & 56 & 61 & \textbf{80} \\
\bottomrule
\\
\end{tabular}
\caption{Data verification results. The in-context learning scores show how well models can infer our tasks, and the human study scores show how often humans write the correct instruction given the instruction induction prompt. All scores above or equal to 80\% are in bold.}
\label{tab:verification_tab}
\end{table}